# 3D-SiamRPN: An End-to-End Learning Method for Real-Time 3D Single Object Tracking Using Raw Point Cloud

Zheng Fang, *Member, IEEE*, Sifan Zhou, Yubo Cui, and Sebastian Scherer, *Senior Member, IEEE*

*Abstract*—3D single object tracking is a key issue for autonomous following robot, where the robot should robustly track and accurately localize the target for efficient following. In this paper, we propose a 3D tracking method called 3D-SiamRPN Network to track a single target object by using raw 3D point cloud data. The proposed network consists of two subnetworks. The first subnetwork is feature embedding subnetwork which is used for point cloud feature extraction and fusion. In this subnetwork, we first use PointNet++ to extract features of point cloud from template and search branches. Then, to fuse the information of features in the two branches and obtain their similarity, we propose two cross correlation modules, named Pointcloud-wise and Point-wise respectively. The second subnetwork is region proposal network(RPN), which is used to get the final 3D bounding box of the target object based on the fusion feature from cross correlation modules. In this subnetwork, we utilize the regression and classification branches of a region proposal subnetwork to obtain proposals and scores, thus get the final 3D bounding box of the target object. Experimental results on KITTI dataset show that our method has a competitive performance in both Success and Precision compared to the state-of-the-art methods, and could run in real-time at 20.8 FPS. Additionally, experimental results on H3D dataset demonstrate that our method also has good generalization ability and could achieve good tracking performance in a new scene without re-training.

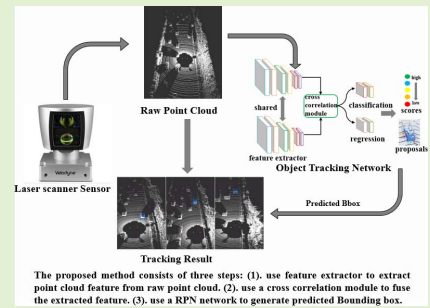

The proposed method consists of three steps: (1). use feature extractor to extract point cloud feature from raw point cloud. (2). use a cross correlation module to fuse the extracted feature. (3). use a RPN network to generate predicted Bounding box.

## I. INTRODUCTION

OBJECT tracking, which is a key issue in computer vision and robotics, could be used in a wide range of applications, such as augmented reality, self-driving cars and mobile robotics. Traditionally, cameras are widely used for object tracking, known as *visual object tracking (VOT)*. For mobile robots or self-driving cars, *3D object tracking* is usually more important than 2D visual tracking. Especially for autonomous vehicles, mobile robots and unmanned aerial vehicles, they usually need not only to track the targets robustly, but also localize (position and orientation) the target objects accurately for efficient path planning, obstacle avoidance and autonomous following, etc. Specifically, *3D single object tracking (SOT)* is of great importance for autonomous following robots or autonomous drones. For example, for an autonomous person following robot, the robot should robustly track its master and localize him/her accurately for efficient following control in the crowd. Another example is autonomous landing of unmanned aerial vehicles, where the drone need to lock the target and know the accurate distance and pose of the target for safe landing.

Existing systems are usually equipped with RGB cameras to track target objects using 2D images. Cameras have many advantages for tracking problems, such as they are compact and cheap, and they can also provide abundant information. However, 2D visual tracking also has its own limitations in practice, especially for autonomous person following robots or autonomous drones. For example, visual tracking usually could not work robustly in visual degraded or illumination changing environments which person following robots or autonomous

Manuscript received September 6, 2020; accepted October 19, 2020. Date of publication October 22, 2020; date of current version January 15, 2021. This work was supported in part by the National Natural Science Foundation of China under Grant 62073066; in part by the Science and Technology on Near-Surface Detection Laboratory under Grant 6142414200208; in part by the Fundamental Research Funds for the Central Universities under Grant N182608003 and Grant N172608005; and in part by the Major Special Science and Technology Project of Liaoning Province under Grant 2019JH1/10100026. The associate editor coordinating the review of this article and approving it for publication was Prof. Chao Tan. *(Corresponding author: Zheng Fang.)*

Zheng Fang, Sifan Zhou, and Yubo Cui are with the Faculty of Robot Science and Engineering, Northeastern University, Shenyang 110819, China (e-mail: fangzheng@mail.neu.edu.cn).

Sebastian Scherer is with the Robotics Institute, Carnegie Mellon University, Pittsburgh, PA 15213 USA.

drones always suffer. Second, visual tracking usually provides pixel coordinate of the tracking target, but lacks accurate target distance information which is important for obstacle avoidance or path following control.

In addition to visual sensors, nowadays 3D LIDAR sensors are also widely used in mobile robots or drones. Compared to 2D images, 3D point cloud generated by laser scanners has more accurate distance measurement and is more robust against illumination change. For this reason, 3D point cloud is also widely used for object or person tracking. However, in contrast to 2D image processing, the processing of point cloud data has its own challenges. *First*, point cloud data is unordered. For example, a point cloud containing N points has N! permutations to represent, and different representation leads to different feature extraction, which makes it hard for neural networks to learn point cloud features end-to-end. *Second*, nowadays point cloud generated by 3D laser scanners is usually much sparser than 2D image, especially for objects at long distances. Sparse point cloud carries little environmental information compared to dense 2D image, which makes it difficult to extract features from point cloud while feature extraction is usually a key step in the tracking problem. *Third*, 3D object tracking needs to estimate not only the position, but also the size and orientation of the target object. The search space dimension (e.g. $x, y, z, w, h, l, ry$) is much higher than 2D visual tracking, which brings great challenges for real-time tracking. *Furthermore*, compared with rigid objects tracking (such as cars), person tracking using sparse point cloud is more challenging since it is more difficult to extract stable features with the non-rigid characteristics.

Currently, most existing 3D object tracking methods are mainly based on the *tracking-by-detection* framework. In this framework, tracking methods are usually composed of two interleaved steps: (i) Target detection: classify point cloud clusters based on handcrafted features [1]–[3] or features extracted by CNN [4], [5]. and (ii) Target tracking: determine the most likely cluster that may be the target object by using filtering-based tracker [6]–[8]. For methods based on such framework, there are some shortcomings. For example, those methods usually have to classify point cloud clusters for every frame using the target detector, which increases the computation burden. Meanwhile, the trackers have to take the output of the detector as their input, which means they cannot track a target if the detector fails to detect the target. Third, a good object tracking method should be class-agnostic, but those methods are usually limited to the specified kind of objects that the detector could classify. In practice, these problems limit the performance of the tracking methods. In recent years, there are also some methods [9]–[12] trying to focus on the tracking ability to avoid the problems of the *tracking-by-detection* framework. For example, a 3D object tracking network is proposed in [9] by mainly utilizing a 3D Siamese tracking network with a Shape Completion network. However, even on a high performance computer with Nvidia GTX1080Ti, the method still could not run in real-time (only 1.8 FPS in our test). Therefore, it is difficult to be deployed on small robot systems, like person following robots or drones. Cui *et al.* [10] also proposed a Point Siamese Network for person tracking. However, the method only predicts the position (x,y,z coordinates) of the target but no orientation and size information are predicted. Besides, Zarzar *et al.* [11] used a 2D Region Proposal Network(RPN) with Bird Eye View(BEV) representation of LIDAR point clouds to generate a small number of object proposals for 3D object tracking. However, the method results in the loss of fine-grained shape information of point cloud due to the BEV representation and does not take into consideration the shape deformations that occur when a person is walking. Recently, Qi *et al.* [12] adopted a Siamese Network to tackle 3D object tracking based on VoteNet [13]. However, their method has similar problems to [10], which could not estimate the size information of objects that are important in real-scene applications.

In this paper, we propose a 3D tracking method called 3D-SiamRPN Network which could directly output the 3D bounding box of the target object in real-time by using raw 3D point cloud. The proposed network consists of two subnetworks. One is feature embedding subnetwork, the other one is region proposal subnetwork. The feature embedding subnetwork has two input branches to extract features from template and search point cloud respectively. In this subnetwork, we first use PointNet++ [14] to extract features of point cloud from both branches. Then, the extracted features from the two branches are input into a feature fusion module which is the core of the subnetwork to fuse feature information. The feature fusion module needs not only to fuse the information of the two groups of features but also to represent the similarity of them. For this reason, we propose two types of cross correlation modules, named Pointcloud-wise and Point-wise respectively. The second subnetwork is a region proposal network (RPN). The correlation feature, which is the output of the cross correlation module, could also be considered as a weight map for the search feature. Therefore, we input the product of correlation feature and the search feature to the region proposal subnetwork to get the proposals and their corresponding scores. Furthermore, we utilize the bin-based 3D box regression loss for the final proposal generation to have a more precise prediction. The proposed method achieved a good balance between speed and accuracy for real-time applications. Experimental results on KITTI dataset show that our method has a competitive performance in both Success and Precision compared to the state-of-the-art methods, and could run in real-time at 20.8 FPS. Additionally, experimental results on H3D dataset demonstrate that our method also has good generalization ability and could achieve good tracking performance in a new scene without re-training.

The main contributions of this paper are as follows:
- We propose a point cloud-based network for the task of 3D object tracking. The proposed network takes siamese architecture and utilizes region proposal subnetwork to track target object. Meanwhile, we use a search area mechanism to avoid our method heavily depending on the front-end detector. The detector is only activated in the first frame or when the target is totally lost, which makes our method more computational efficient.

- To the best of our knowledge, this is the first work to propose cross correlation modules on 3D point cloud data for 3D object tracking tasks. Our proposed Pointcloud-wise and Point-wise cross correlation modules are very simple and effective. Experimental results show that they have better performance compared to other similarity measurement modules.
- We carried out quantitative and qualitative experiments to validate the performance of proposed method. Experimental results on KITTI dataset show that our proposed method has a state-of-the-art performance in most metrics with real-time speed. Meanwhile, the experiment on H3D dataset also shows the good generalization ability of our method.

The rest of this paper is organized as follows. In section II, we discuss the related work. Section III describes the proposed 3D-SiamRPN method. We validate the performance of our methods on KITTI and H3D datasets in section IV and we conclude in section V.

## II. RELATED WORK

Currently, object tracking could be mainly divided into 2D visual tracking and 3D object tracking depends what kind of sensors are used. In this section, we have a brief overview of those two kinds of tracking methods.

### A. 2D Visual Tracking

Early works in 2D visual tracking were mainly based on Correlation Filters (CF) [15]–[20]. For example, Bolme *et al.* [15] introduced CF to object tracking and achieved high tracking speed, but the tracking accuracy is not good enough. After that, Henriques *et al.* [16] proposed a new Kernelized Correlation Filter tracker based on Histogram of Oriented Gradients (HOG) features instead of raw pixel and achieved a better performance. Bertinetto [17] combined two image patch representations that are sensitive to complementary factors to improve the robustness of the tracker against both color changes and deformations. However, they are both sensitive to scale changing. Danelljan *et al.* [18] solved this by using 33 different scales. After that, Danelljan *et al.* [19], [20] used deep features and an interpolation model which highly improved the performance of the tracker. However, these methods could not cope with fast deformation well since they are based on template matching.

Recently, with the advance of deep learning, methods based on Siamese network [21] become the main framework in visual tracking. The pioneering work using this architecture is the fully convolutional siamese network (SiamFC) [22]. In that paper, it is the first time that researchers treated visual tracking as a similarity issue and introduced the correlation layer into the architecture. They also integrated correlation filter [23] into a layer in CNN network later. The two networks have good performance in VOT-2014 tracking benchmark [24]. However, compared to CF methods, their performance of accuracy and robustness are unsatisfying. Meanwhile, their method is not suitable for noisy backgrounds and crowded scenes. Different from [23], Guo *et al.* [25] adopted videos as input instead of image pairs in siamese network, which could learn the consistency of the appearance and background of the object online. Meanwhile, they made up for the shortcoming of the update process ignored by the tracker based on the siamese network, so that the overall accuracy of the network has been greatly improved. But their method still has some performance gap compared with CF methods. Afterwards, Li *et al.* [26] introduced the RPN to Siamese network. SiamRPN could regress a preciser 2D bounding box for the target object and outperform the state-of-the-art methods based on CF. However, the method does not cope well with disturbances. Based on [26], Zhu *et al.* [27] introduced some other datasets to balance the training samples and augmented negative samples to improve the discriminating ability of tracker. However, the basic architecture of network is still restricted at AlexNet which could not utilize high-level semantic features. Li *et al.* [28] broke the restriction and excavated the ability of deep network. They also proposed a depth-wise layer to do cross correlation operation and achieved state-of-the-art performance on VOT-2018 [29] for short-term video object tracking. Based on [28], Wang *et al.* [30] produced class-agnostic object segmentation masks and rotation bounding boxes estimation. However, their estimation of segmentation mask is not accurate enough. Recently, Voigtlaender *et al.* [31] proposed a Siamese two-stage detection network for visual tracking, which take advantage of re-detection of the first-frame template and previous-frame, to model the object to be tracked. Their method outperformed all previous methods on six short-term tracking benchmarks as well as on four long-term tracking benchmarks, where it achieves especially strong results. In summary, 2D visual tracking methods have achieved great progress in the past decade and have been applied to many practical scenarios. However, 2D visual tracking methods are still sensitive to illumination change. Additionally, most 2D visual tracking methods only obtain pixel coordinate of the tracking target, however sometimes we need to know the accurate 3D pose the tracking target.

### B. 3D Object Tracking

Previous works usually use *tracking-by-detection* framework to track 3D objects. They first utilize an off-the-shelf detector [32], [33] to detect objects and then use probability methods [6], [34]–[39] to match the detection results overtime. These tracking methods can be divided into two categories according to the detectors used, one-stage(single-shot) detector-based tracking methods [36], [37] and two-stage detector-based tracking methods [38], [39]. Simon *et al.* [36] used a modified one-stage detector [40] to detect target and used Labelled Multi-Bernoulli Filter to track the target with 100FPS. Luo *et al.* [37] proposed a one-stage detector and used the overlap between the detection results to track objects with 33 FPS. However, [37] mainly focused on the detector and need to convert point cloud to Bird's Eye Views (BEV), which may lose information of point cloud. For two-stage detector-based tracking methods, Shenoi *et al.* [38] adopted a two-stage detector [33] to obtain 3D detection and used Joint Probabilistic Data Association(JPDA) to perform the data matching for tracking with 14FPS. Baser et al. [39] first used a two-stage detector [32] and then proposed a matching net to compare detection results with 25 FPS. Generally, the one-stage detection tends to be faster and simpler, while the two-stage detection tends to achieve higher precision. Benefiting from the more and more accurate detectors, they could obtain good performance with simple trackers, but different detectors will result in different tracking results. However, those methods usually pay much attention to the detection or data association part but neglect the tracking part. And, they usually have high computation cost since they need to detect objects in every frame. Meanwhile, it could not track 3D objects in an end-to-end manner due to the separation of the detector and tracker. Recently, with the advance of point cloud representation based on deep learning [14], [41], [42], some researchers try to solve this problem based on neural networks. Frossard and Urtasun [43] proposed a neural network to detect and track 3D object in an end-to-end manner. Scheidegger et al. [44] also implemented an improved network to detect object and applied PMBM tracking filter to associate object with a mono-camera. Hu et al. [45] used a LSTM network to associate 3D vehicle detection results over time based on a mono-camera to achieve vehicle tracking. However, these methods both need RGB image information and rely on object detector to provide prior information. Recently, some methods try to solve the tracking problem in an end-to-end manner based on the siamese networks. Giancola et al. [9] utilized a 3D Siamese network to track 3D object. Meanwhile, they used cosine similarity as their cross correlation technique to represent the similarity of the two point clouds. However, they still need a front-end search method to provide search candidates as their input every frame and could not run in real-time. To efficiently search for the target object, Zarzar et al. [11] leveraged a 2D Siamese network to generate a large number of coarse object candidates on BEV representation. However, the method is not able to identify a correct ranking over the other top K candidates, resulting in the inability to select the optimal candidate. Besides, they did not focus on more effective cross-correlation methods to fuse template and search features. Similarly, Qi and Feng [12] used a Siamese Network to solve 3D object tracking based on VoteNet [13]. They first fused the template and search seeds with a specific approach, then they used VoteNet to generate potential object centers(votes) and estimated position and orientation of the target center based on those votes. In their feature fusion approach, they first calculated the cosine similarity between the template and search seeds, and then concatenated template seeds and the similarity matrix. Finally, the concatenated features are input into Multi-layer perceptron to obtain search area seeds with target-specific features. However, their method could not regress the size information of objects and their performance needs to be improved. In this paper, we aim to improve the predictive ability of the back-end tracking for single object tracking. We propose a 3D-SiamRPN network to solve 3D single object real-time tracking problem by using raw 3D point cloud in an end-to-end manner.

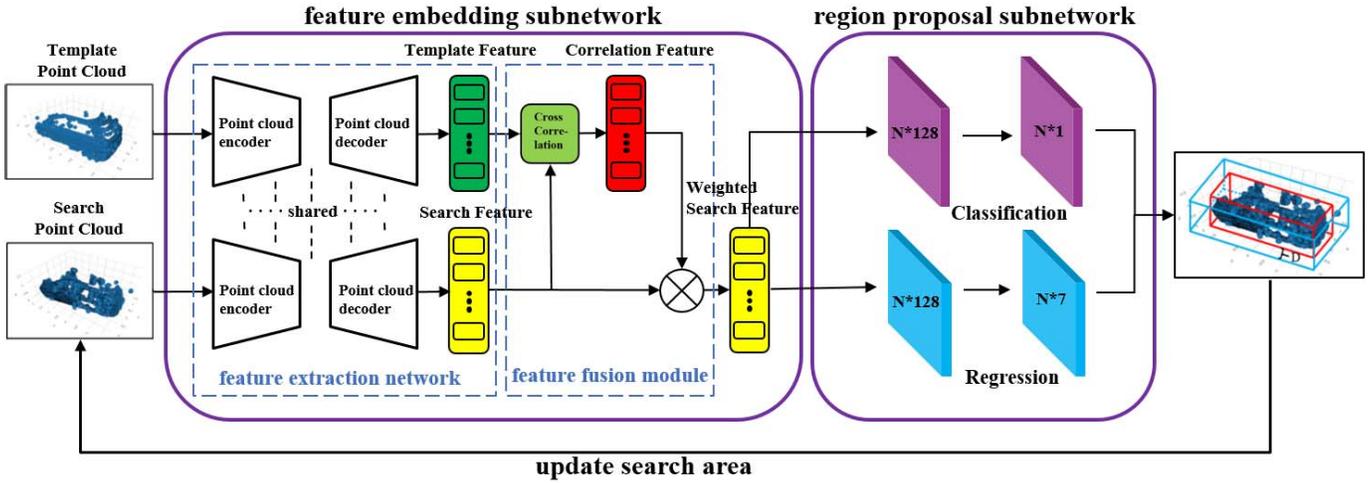

Fig. 1. The architecture of the proposed network. It consists of two subnetworks: feature embedding subnetwork and region proposal subnetwork. The rightmost red box is the predicted box and the blue one is the search area. And in the next frame, all point clouds in the search area will be the search input for the network. Therefore, we could form a closed loop to track object directly.

## III. METHOD

The overall architecture of our 3D-SiamRPN network is depicted in Fig. 1. The proposed network is composed of a feature embedding subnetwork and a region proposal subnetwork. In inference stage, 3D-SiamRPN not only predicts 3D bounding box of the target object, but also obtains a predicted search area where the target object may appear in the next point cloud frame. The final search area is obtained by applying an additional range $D$ to the predicted bounding box of target object. After tracking completed for one frame, the point cloud in the search area in the current frame will be used as the search input for the network in the next frame. We introduce each subnetwork in detail in the remainder of this section.

## A. Feature Embedding Subnetwork

The feature embedding subnetwork includes a feature extraction network and a cross correlation module. The feature extraction network takes Siamese architecture to extract features from the input point clouds. Then, the cross correlation module obtains a similarity map between extracted features in the high-level feature space. The feature extraction network consists of two branches called template and search branches respectively. The input of the template branch is the point cloud of the initialized target object (could be obtained by reading label), and the input of the search branch is the point cloud within the search area in the current frame. The two inputs are denoted by $Z$ and $X$ separately. It should be noted that the template branch $Z$ will be frozen and stop updating after initialization, thus we only update the search branch $X$ in each frame of point cloud. In feature extraction step, the inputs of two branches are re-sampled to the same size firstly, and then encoded with the feature extraction network to obtain point cloud features of the template and search branches. Hereafter, we call them *template feature* and *search feature*. The two branches of the networks shares parameters in the point cloud encoder-decoder network so that the two inputs are implicitly encoded by the same transformation, which is suitable for the subsequent tasks. After the feature extraction network, we propose two different types of cross correlation modules to embed the two features and obtain a *correlation feature* representing their similarity. Meanwhile, the correlation feature could be considered as a weight map for the search feature. Thus the final output of feature embedding subnetwork is the product of search feature and correlation feature. We denote $\varphi(z)$ and $\varphi(x)$ as the output feature maps of the feature extraction network, and denote $\psi$ and $\varphi(x)'$ as the correlation feature and the final weighted search feature respectively.

*1) Feature Extraction Network:* In order to achieve excellent tracking performance, we need to extract stable and distinctive features from 3D point cloud. There are many point cloud features proposed in the past decade. However, most existing point cloud features are handcrafted towards specific tasks, such as PFH, FPFH, VFH, ESF, etc [46]. Tracking methods based on such kind of features only perform well on specific object. In order to achieve class-agnostic tracking ability using raw point cloud, in this paper, we use PointNet++ [14] with multi-scale grouping to extract point cloud features. The method first uses a mini-network(T-Net) to align all points to a canonical space before feature extraction, and then uses another T-Net to align feature space established by the Multilayer Perceptron(MLP), which are called input transform and feature transform respectively. Finally, it uses MLPs to extract high-level features. This method not only preserves the integrity of point cloud data mostly but also extracts invariant features with unordered raw point cloud data. Besides, it also exploited the sparse and local structure of the point cloud to further improve the quality of feature extraction.

*2) Cross Correlation Module:* Cross correlation module fuses the two sources of features obtained from the feature extraction network, which is the core operation in feature embedding subnetwork. Cross correlation module has been used in many ways, such as feature fusion and measurement of similarity. In 2D visual tracking, SiamFC [22] first uses a cross correlation layer to obtain a response map indicating the target location. SiamRPN [26] extends the cross correlation layer to embed higher-level feature information and outputs anchors and classifications. Furthermore, SiamRPN++ [28] proposes a lightweight cross correlation layer, called depth-wise cross correlation, which has fewer parameters but achieves better performance. In 3D object tracking, SC3D [9] use a simple cosine distance to represent the similarity between two pairs vectors. P2B [12] also use the same cosine similarity as SC3D to indicate which part of the search point cloud is more similar to the template point cloud. However, the cosine distance only measures the cosine value of angle between two vectors. Although it is suitable for images, it is not necessarily suitable for point cloud data. As far as we know that there is almost no method using cross correlation module for point cloud object tracking yet.

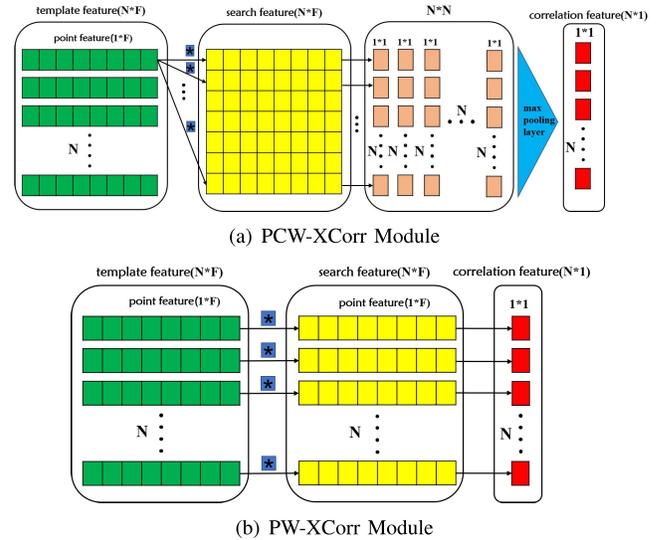

Fig. 2. The two cross correlation modules. (a) The PCW-XCorr module (b) The PW-XCorr module. The green one is template feature $\varphi(z)$ and yellow one is search feature $\varphi(x)$, the final correlation feature $\psi$ is in red. $*$ means correlation operation.

In this paper, we present two types of cross correlation modules, named Pointcloud-wise Cross Correlation (PCW-XCorr) and Point-wise Cross Correlation(PW-XCorr) respectively, as shown in Fig. 2. The two modules can both fuse the information of features and obtain a similarity feature. The main difference between them is how the features are convolved. In the PCW-Xcorr module, the $N \times F$ size template feature $\varphi(z)$ is first divided into $N$ individual $1 \times F$ size point features. Each point feature is considered to be a kernel to do correlation operation on the $N \times F$ size search feature $\varphi(x)$. Therefore, for the search feature, N correlation operations by a $1 \times F$ kernel will produce a feature map with the size of $N \times N$. After all cross correlation operations, we concatenate all of these features and apply a max pooling layer to obtain a $N \times 1$ size correlation feature.

However, the PCW-XCorr module focuses more on the global similarity between two groups of features (template feature and search feature), and it is computational demanding since the entire search feature is convolved every time. Therefore, we propose an alternative approach that focuses more on local similarity and avoid heavy computation. Different from PCW-XCorr module, PW-XCorr divides both $\varphi(z)$ and $\varphi(x)$ into $N$ individual $1 \times F$ size point features to do cross correlation operation point by point. It should be noted that the features of template and search point cloud extracted by PointNet++ have the same transformation since the two branches of the feature extraction network share network parameters and have already been aligned in high-level space using a T-Net. Therefore, the order of points in raw point cloud has no influences on the cross correlation operation. After the cross correlation operation, we could obtain a $1 \times 1$ feature for each point which indicate the local similarities. These local similarities could constitute the global similarity, thus we concatenate all $1 \times 1$ output features to obtain a $N \times 1$ size correlation feature $\psi$.

The two modules could be formulated as follows:

$$\psi = \max(\sum_{i \in N} \varphi(z_i) \otimes \varphi(x)) \quad (1)$$

$$\psi = \sum_{i \in N} \varphi(z_i) \otimes \varphi(x_i) \quad (2)$$

where $\otimes$ means cross correlation operation, $N$ is the number of the points in point cloud, $\varphi(z_i)$ and $\varphi(x_i)$ are corresponding point features of $i^{th}$ point in $\varphi(z)$ and $\varphi(x)$ respectively.

By applying the two types of cross correlation modules, we can obtain a correlation feature $\psi$ to describe the similarity between features from two branches. Specifically, if a point in search feature is more similar with a point in template feature in PW-XCorr module, its corresponding value in $\psi$ will be higher. Meanwhile, if a point in search feature is more similar with the total template feature in PCW-XCorr module, its corresponding value in $\psi$ will be higher. Therefore, the correlation feature $\psi$ could also be considered as a weighted feature for the search feature. For this reason, we multiply $\psi$ with the search features $\varphi(x)$ to obtain the final weighted search feature $\varphi(x)'$, which will be the input of the next subnetwork. The distribution of $\varphi(x)'$ becomes similar to that of $\varphi(z)$ after weighting. The more similar features are up-weighted and the others are down-weighted. An example of this process is shown in Fig. 3.

### B. Region Proposal Subnetwork

In order to obtain more accurate target position prediction, the tracker needs to predict the 3D bounding box of the target object. The 3D bounding box could be represented as *(x, y, z, w, h, l, ry)* in LIDAR coordinate, where *(x, y, z)* is the location of object center, *(w, h, l)* is the size of object, *ry* is the heading angle of the object from BEV. Therefore, the tracker should accurately predict these 7 values for successful tracking. It is obvious that the search dimension of 3D object tracking is much higher than 2D tracking, which poses great challenges for real-time tracking. In order to generate predicted box based on the fusion features, we consider to use 3D detector methods to regress 3D bounding box. As mentioned in [47], one-stage detection methods do not need region proposal generation and post-precessing, such as VoxelNet [48], SECOND [49]. Two-stage detection methods first propose coarse regions that may include objects and then estimate bounding boxes, such as PointRCNN [50], FVNet [51]. The one-stage detection tends to be faster and simpler, while the two-stage detection tends to achieve higher precision. In this paper, we not only focus on the high running speed for realistic application, but also need to pay more attention to accurate performance, to achieve the balance between speed and accuracy for real-scene application. Therefore, we use a modified Region Proposal Network(RPN) proposed in PointRCNN [50] to obtain the final 3D bounding box of the target object. In that work, points inside the 3D label boxes are considered as foreground points and others are background points. Each point generates one proposal to avoid using too many anchor boxes. In our work, in order to reduce the computational burden, we input the RPN with the final weighted search feature $\varphi(x)'$ to generate proposals and scores. The region proposal subnetwork consists of a classification branch and a proposal generation branch. The classification branch is for foreground-background point classification and the regression branch is for bin-based 3D proposal generation. We introduce these two parts in detail in the remainder of this section.

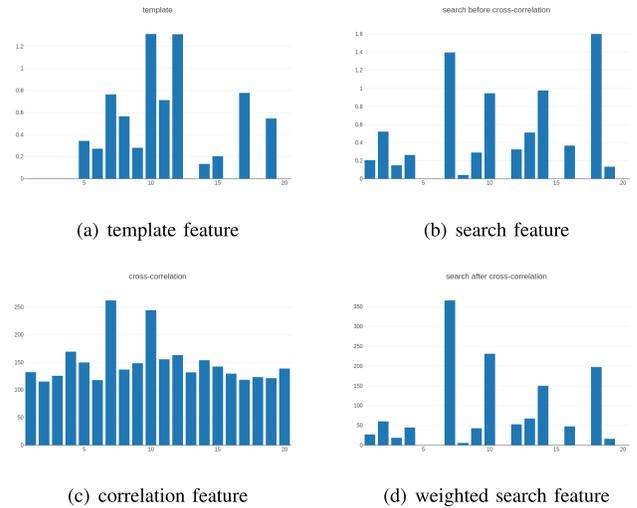

Fig. 3. Feature examples. (a) The template feature $\varphi(z)$. (b) The original extracted search feature $\varphi(x)$. (c) The correlation feature $\psi$. (d) The final weighted search feature $\varphi(x)'$. After multiply with the cross correlation feature, the distribution of the search feature is similar to that of the template feature.

*1) Foreground Point Classification:* The foreground points provide rich information about the location and direction of its associated objects. By distinguishing the point cloud into foreground and background points, a small number of high-quality 3D proposals could be generated directly rather than using a large number of redundant 3D anchor boxes. Meanwhile, we could effectively constrain the search space for 3D proposal generation. In the foreground point classification block, we use two 1D-convolutional layers with filter size [128,1] to get the point-wise classification results based on the weighted search feature $\varphi(x)'$. We assign a binary class label (of being foreground point or not) to each point. We assign a positive (set as 1) label to the point inside the 3D label boxes, i.e. foreground points, and a negative (set as -1) label to the point that is not in the 3D label boxes, i.e. background points. The number of foreground points is usually much smaller than that of background points in large-scale outdoor scene. Therefore, we use focal length loss [52] to balance the two classes.

$$L_{focal}(p_t) = -\alpha_t(1-p_t)^\gamma \log(p_t) \quad (3)$$

$$\text{where } p_t = \begin{cases} p & \text{if } label = 1 \\ 1-p & \text{otherwise} \end{cases} \quad (4)$$

where $p \in [0, 1]$ is the points' estimated probability for the class with label $= 1$, $p_t$ is the classification probability of different classes, $\alpha_t \in [0, 1]$ is a weighting factor for addressing class imbalance, $\gamma \in [0, 5]$ is a tunable focusing parameter for reducing the relative loss for well-classified examples($p_t > .5$). During the point classification training, we set $\alpha_t = 0.25$ and $\gamma = 2$ as the original paper.

*2) Bin-Based 3D Proposal Generation:* In this section, we first generate bin-based 3D object proposal boxes, and then regress object location based on these proposals. Note that the proposal box generation is done as the foreground point classification at the same time. After generating proposal boxes, we need to regress it to obtain the accurate object location. In regression block, we use two 1D-convolutional layers with filter size [128,7] to get the point-wise regression results based on the weighted search feature $\varphi(x)'$. The dimension of the network regression result is $B \times N \times C$, where B is data batch size, N is the number of point cloud, and C is determined by the partition number of *(x, y, z, w, h, l, ry)*, which we would mention it below. To have a more accurate bounding box prediction, we use the bin-based regression loss proposed in [50] to estimate the coordinates of target object. The main idea of bin-based regression loss is to subdivide the region around the foreground points along each coordinate axes. The search area is divided into a series of bins along the *X, Y, Z* axes. Specially, a search range *S* is set for each current foreground point along each axis, and is divided into bins of uniform length *l*. Similarly, the orientation $2\pi$ is also divided into *n* bins. For the bin-based regression, the *reg* layer would estimate a confidence probability and a residual value. The probability is utilized to obtain which bin the point belongs to and the residual is added to refine the final values. Fig. 4 shows the bin-based bounding box localization. To estimate the size of the target object $(w, h, l)$, we directly regress their residual value since their values are within a small range.

The localization loss consists of two parts, one for bin classification of $x, y, z, \theta$, and the other for residual regression based on bin classification. In the process of locating the bounding box of the target object, the location is estimated roughly by bin classification first, and then the result of residual regression is added to the bin classification to realize further localization. Therefore, the location of target object could be formulated as following [50]:

$$bin_u^i = \frac{u^t - u^i + S}{l} \quad (5)$$

$$res_u^i = \frac{1}{C}(u^t - u^i + S - (bin_u^i * l + \frac{l}{2})) \quad (6)$$

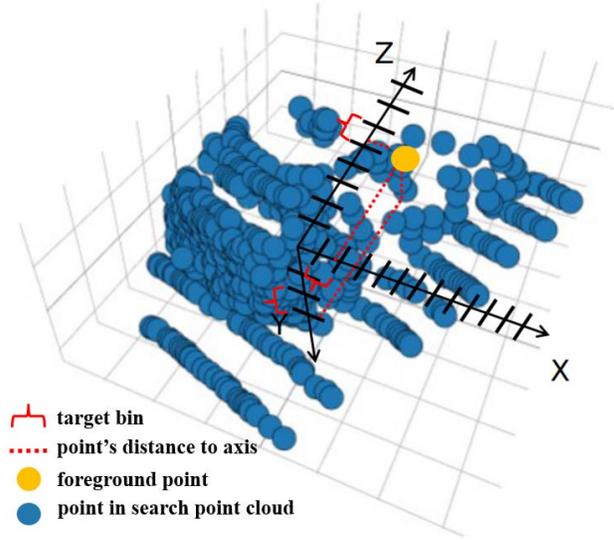

Fig. 4. The example of bin-based bounding box localization. The surrounding area along every coordinate axes of each foreground point is divided into a series of bins to locate the object center.

where $u \in x, y, z, \theta$, $u^t$ is the center coordinates of the target object, $u^i$ is the coordinate of the input $i^{th}$ point. $l$ is the uniform length of bins divided by the search range $S$. $bin_u^i$ is the ground-truth bin classification along $X$, $Y$ and $Z$ axis, $res_u^i$ is the refined ground-truth residual for further localizing the target bounding box, $C$ is the bin length for normalization.

*C. Loss Function*

The overall loss includes classification and regression terms. We use the focal loss in classification branch. Meanwhile, we follow the regression loss of [50] which includes the bin classification and residual regression terms. The regression loss $L_{reg}$ could be formulated as follows:

$$L_{bin}^i = \sum(F_{cls}(\bar{bin}_u^i, bin_u^i) + F_{reg}(\bar{res}_u^i, res_u^i)) \quad (7)$$

$$L_{res}^i = \sum F_{reg}(\bar{res}_v^i, res_v^i) \quad (8)$$

$$L_{reg} = \frac{1}{N_{pos}} \sum (L_{res}^i + L_{bin}^i) \quad (9)$$

where $u \in x, y, z,$, $v \in w, h, l$, $N_{pos}$ is the number of positive samples, $\bar{bin}_u^i$ and $\bar{res}_u^i$ are the predicted bin classification and residuals corresponding to the foreground point p, $bin_u^i$ and $res_u^i$ are the ground-truth targets calculated as above. We use the smooth L1 loss as $F_{reg}$ and the cross-entropy loss as $F_{cls}$. The training loss could be formulated as:

$$L = \sum L_{cls} + \lambda \sum L_{reg} \quad (10)$$

where $\lambda$ is the balance weight for $L_{reg}$.

IV. EXPERIMENTS

We evaluate our 3D-SiamRPN method on challenging 3D object tracking benchmark of KITTI [53] dataset and Honda Research Institute 3D Dataset (H3D) [54] dataset. We first carried out rigid object (car) tracking and non-rigid object (pedestrian) tracking experiments on KITTI dataset.

5002                                                                                                                                IEEE SENSORS JOURNAL, VOL. 21, NO. 4, FEBRUARY 15, 2021Then, we tested our method (trained on KITTI data) on H3D dataset without re-training. Experimental results show that the proposed method has good class-agnostic tracking ability and generalization ability. The results also show that our method could achieve a competitive performance in both Success and Precision compared to the state-of-the-art methods while running in real-time around 20.8 FPS. Our experiment video is available at https://youtu.be/xy6Dh2LseRQ.

### A. Dataset

*1) KITTI:* We use the training set of KITTI tracking benchmark in our experiments. It includes more than 20,000 manually labeled 3D objects captured in cluttered scenes using Velodyne HDL-64E 3D LiDAR (10 HZ). For all following experiments, we focus on vehicle and person tracking since they are the main classes in the KITTI dataset. Besides, they also represent rigid object and non-rigid object respectively. We split the sequences of KITTI following [9]: Sequences 0-16 and 17-18 are used for training and validation respectively in training stage. Sequences 19-20 are used for testing in the inference stage.

*2) H3D:* Recently, Patil *et al.* [54] presented H3D dataset to provide sufficient data and labels to tackle challenging scenes where highly interactive and occluded traffic participants are presented. H3D comprises of 160 crowded and highly interactive traffic scenes with a total of 1 million labeled instances in 27,721 frames. Unlike KITTI dataset where 3D object annotations are only labeled in the frontal view, 3D objects in the h3d dataset are labeled in the full-surround view, as shown in Fig. 16. Therefore, the target object may disappear for some frames and appear again later. Meanwhile, the H3D dataset is collected from crowded urban scenes and is more complex than KITTI dataset. Thus, it is more difficult than KITTI dataset for 3D object tracking. In our experiments, we test our proposed method on scenario 011, 018, 022, 045 and 050 sequences in H3D dataset, which includes more than 120 tracking instances.

### B. Implementation Details

*1) Network Architecture:* For the feature extraction network, we use four set-abstraction layers with multi-scale grouping to sub-sample points into groups with sizes 500, 318, 256, 64, and utilize four feature propagation layers to obtain point-wise feature. Notably, the template point cloud feature is frozen after initialization and would not be updated. In the region proposal subnetwork, we use two one-dimensional convolutional layers for both *cls* and *reg* layers to predict the foreground-background point classification and the target object location. In the classification branch, we employ two convolutional layers sequentially as Conv1D(128,128,1,1,0) and Conv1D(128,1,1,1,0). In the regression branch, we use two convolutional layers sequentially as Conv1D(128,128,1,1,0) and Conv1D(128,7,1,1,0). In order to prevent over-fitting of the model, we insert a Dropout layer($p = 0.5$) between the two convolution layers of both branches.

*2) Training Detail:* In the training stage, the training point cloud data is obtained by reading the bounding box ground truth(GT) of target objects in all sequences. Meanwhile, in order to improve the robustness of the network, the bounding box are enlarged to bring in some noises. We randomly select a pair of point clouds training from point cloud data as template point cloud and search point cloud. Then we randomly re-sample both template and search point clouds to a fixed size, which could be used as the network input. For a point cloud including $N$ points, the bin-based RPN generates $N$ proposals. We use all of the proposals which the corresponding class belongs to the foreground point. The number of points on network input is $N = 500$. We use fixed-size anchors according to the mean sizes of all ground truths in the KITTI training set. Anchor is considered as positive if its corresponding point is the foreground point. We follow the setting of the bin-based proposal generation [50], the search range $S$ along $X$, $Z$ axes are set to $3m$, $S$ along $Y$ axis is set to $0.5m$, the bin size $l$ along $X$, $Z$ axes are set to $0.5m$ while along $Y$ axis is $0.25m$. The number of orientation bin $n$ is 12. The balance weight $\lambda$ is set to 10. In training setting, we use the Adam optimizer with an initial learning rate of 0.002 and a batch size of 10 for 50 epochs. All training process are implemented with a Nvidia 1080ti GPU and Pytorch. In the tracking experiments, the proposed method was implemented in ROS Kinetic framework running on Ubuntu16.04 LTS and Intel Core E-2288G CPU. All coordinates are based on our Velodyne coordinate system.

*3) Model Inference:* In the inference stage, we initialize the template branch by reading 3D bounding box GT of first frame of target trajectory. The first search area is obtained by enlarging the bounding box of the first frame, and then we crop the point cloud in this search area from the next frame to input search branch. After model inference, we get 500 proposals with different classification scores. We first select the top 100 proposals with the highest score, then use Non-Maximum Suppression(NMS) with IoU threshold 0.8 to filter those 100 proposals, and use the proposal with the highest score as the final predicted box of the target in the next frame. Finally, the subsequent search areas are updated by enlarging the predicted box. Details of the enlarging search range based on predicted boxes could be found in Fig. 8 and 12.

### C. Evaluation Metrics

We report the *Success* and *Precision* metrics defined by One Pass Evaluation (OPE) [55] for Single Object Tracking, which represent overlap and error Area Under the Curve (AUC) respectively.

### D. 3D Single Rigid Object (Car) Tracking on KITTI

For rigid object (car) tracking, in order to evaluate the performance of our method and discuss the selection of suitable parameters, we carried out 5 different experiments on KITTI dataset. In the first experiment, we qualitatively analyze the performance of our method in several challenging scenes. The challenging scenes include crowded scenes, extremely sparse scenes and turning scenes. In the second experiment, we quantitatively evaluate the performance of our method. We compare the Success and Precision with the

Authorized licensed use limited to: Northeastern University. Downloaded on January 20,2021 at 07:40:55 UTC from IEEE Xplore.  Restrictions apply.

TABLE I
EVALUATION RESULTS OF CAR TRACKING ON KITTI

| Module | Modality | 3D Success | 3D Precision | BEV Success | BEV Precision | FPS |
|---|---|---|---|---|---|---|
| STAPLE CA [17], [56] | RGB | - | - | 31.60 | 29.30 | 35.2 |
| AVOD Tracking [32] | RGB+LIDAR | 63.16 | 69.74 | 67.46 | 69.74 | - |
| SC3D-KF [9] | LIDAR | 40.09 | 56.17 | 48.89 | 60.13 | 2.2 |
| SC3D-EX [9] | LIDAR | **76.94** | **81.38** | **76.86** | **81.37** | 1.8 |
| 3D Siamese-2D proposal(top-16) [11] | LIDAR | 36.3 | 51.0 | - | - | - |
| P2B [12] | LIDAR | 56.2 | 72.8 | 72.9 | 78.4 | **45** |
| Ours(PCW-Xcorr) | LIDAR | 56.32 | 73.40 | 72.21 | 77.19 | 16.7 |
| Ours(PW-Xcorr) | LIDAR | 57.25 | 75.03 | 73.02 | 79.45 | 20.8 |

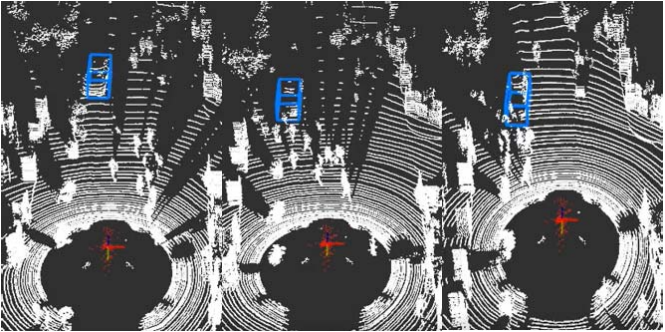

Fig. 5. The example of car tracking in a crowded scene where the target vehicle ID is 42 in sequence 19 of KITTI dataset. The blue box is the target object. In this scenario, approximately 14 pedestrians are moving in an area of $20m^2$ in front of the tracked car.

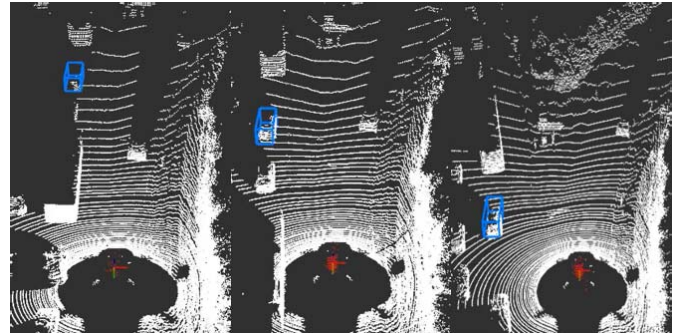

Fig. 7. The example of car tracking in a turning scene where the target car ID is 97 in sequence 20 of KITTI dataset. The blue box is the target object. In this scenario, the angular velocity of LIDAR is approximately $5°/s$.

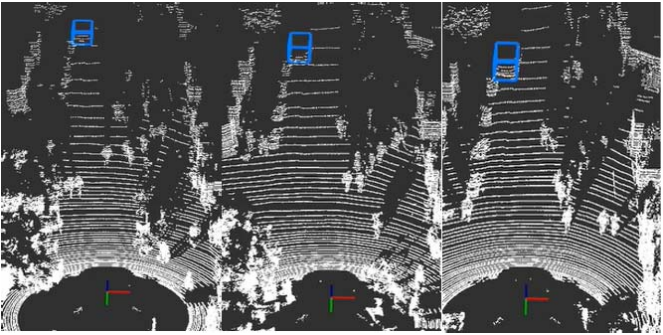

Fig. 6. The example of car tracking in an extremely sparse scene where the target car ID is 42 in sequence 19 of KITTI dataset. The blue box is the target object. In this scenario, when the relative distance is about $52m$, the number of points for target car point clouds is only about 10. When the relative distance is about $40m$, the number of points for target car point clouds is about 36.

state-of-the-art methods. In the third experiment, we quantitatively evaluate the performance differences between our cross-correlation modules (PCW-Xcorr, PW-Xcorr) and other commonly used cross-correlation modules. In the fourth experiment, we test the effect of the search range on the performance of the method, because different search area results in different updated search point clouds. In the fifth experiment, we analyze the effect of different weight coefficient on the tracking performance.

*1) 3D Car Tracking in Challenging Scenes:* The tracking performance of the proposed method for car tracking on the KITTI dataset is shown in Fig. 5, Fig. 6 and Fig. 7, where the blue bounding box is the target car. In Fig. 5, the proposed method can track a car effectively in a crowded pedestrian scene. In this scenario, approximately 14 pedestrians were walking in an area of $20m^2$ in front of the target car. As we can see from Fig. 5, most of the point cloud of the car are occluded by the pedestrians. Since there are many pedestrians around the target car, it will introduce the point cloud of pedestrians into the search point cloud when updating the search point cloud of the car. However, our method could still effectively track the target car. In Fig. 6, our method has good tracking performance even in a sparse scene. In this scenario, the number of points for target car point clouds is only 10 when the relative distance is about $52m$, and the number of points for target car point clouds is only 36 when the relative distance is about $40m$. Even the target car has extremely sparse point clouds, our method could still effectively track it. In Fig. 7, our method could track the target car in a turning scene successfully. In this scenario, the angular velocity of LIDAR is approximately $5°/s$.

*2) Evaluation of 3D Car Tracking:* As shown in Table I, our proposed methods both surpass the SC3D-KF [9], 3D Siamese-2D proposal(top-16) [11] in all metrics. Notably, although SC3D-EX [9] has better performance than other methods in all metrics, their method is unreasonable because it updated search area based on the bounding box ground truth(GT) of current frame. However, it is difficult to obtain the GT of each frame for target object in real scene implementation. Besides, our method outperforms the P2B [12] in all metrics, which is a state-of-the-art method for 3D object tracking. Meanwhile, compared to AVOD Tracking [32] which utilized a RGB+LIDAR detector, our proposed tracker with PW-Xcorr performs better by a large margin of 5.29%, 5.56% and 9.71% in 3D Precision, BEV Success and BEV Precision respectively. It verifies the shortcoming of heavy reliance on the front-end detector of tracking-by-detection methods and proves that the 3D bounding box obtained directly by our tracker could be better than that by RGB+LIDAR detector. More importantly, compared with other methods, our proposed methods could run in real time. This is because we could freeze the template branch without updating it according to the common manner in the existing visual object tracking works [26], [28], [30], and the network will have less computational burden and obtain a faster running speed. The proposed PCW-XCorr and PW-XCorr trackers could achieve 16.7 FPS and 20.8 FPS respectively. Our methods achieve the state-of-the-art performance among the real-time single object tracking algorithms. In other words, our method has a good balance between speed and accuracy.

TABLE II
PERFORMANCE FOR DIFFERENT MODULES OF CAR ON KITTI

| Module | 3D Success | 3D Precision | BEV Success | BEV Precision | FPS |
|---|---|---|---|---|---|
| No similarity Module | 42.62 | 57.43 | 55.99 | 60.88 | **21.4** |
| DW-Xcorr | 52.98 | 70.59 | 71.17 | 75.31 | 19.6 |
| Cosine Similarity | 52.81 | 70.80 | 70.92 | 76.11 | 19.5 |
| Euclidean Similarity | 54.39 | 71.03 | 71.12 | 77.55 | 19.2 |
| Ours(PCW-Xcorr) | 56.32 | 73.40 | 72.21 | 77.19 | 16.7 |
| Ours(PW-Xcorr) | **57.25** | **75.03** | **73.02** | **79.45** | 20.8 |

TABLE III
CAR TRACKING PERFORMANCE FOR DIFFERENT WEIGHTS ON KITTI

| weight($\lambda$) | 3D Success | 3D Precision | BEV Success | BEV Precision |
|---|---|---|---|---|
| 0.1 | 52.48 | 71.51 | 70.77 | 78.13 |
| 0.5 | 51.71 | 71.26 | 71.65 | 78.71 |
| 1.0 | 52.21 | 72.26 | 71.93 | 79.46 |
| 5.0 | 56.46 | 73.02 | 72.60 | 79.63 |
| 10.0 | **57.25** | **75.03** | **73.02** | **79.45** |
| 20.0 | 56.58 | 74.82 | 72.21 | 79.19 |

*3) Cross Correlation Module Ablation Studies for 3D Car Tracking:* In Sec.III-A.2, we introduced two types of cross correlation modules. Here, we compare them with the most widely used cosine and euclidean similarity methods. Meanwhile, we compare them with the depth-wise cross correlation (DW-XCorr) [28], which has a good performance in visual tracking. Additionally, in order to compare the effect of the cross correlation module in our method, we remove the cross correlation module and directly put the search feature into the RPN. Table II reports the results of the comparison. As the results show, the two proposed modules outperform the other three modules in four metrics, and the proposed module PW-XCorr performs better than the others. Meanwhile, compared to other modules, the network without similarity module performs worse by a large margin from 10.19% to 18.57% on Success and Precision metrics. This verifies that it is necessary to take template features into account to get the correlation features. The results show that our proposed modules have better performance than cosine and euclidean similarity modules. Meanwhile, PW-XCorr performs better than DW-XCorr by 4.27%, 4.44%, 1.85% and 4.14% on 3D/BEV Success and Precision respectively. The results also show that our proposed modules are better than DW-XCorr module, and also verify that our proposed modules are more suitable for point cloud tracking. Additionally, compared to other modules, the PW-XCorr module has a faster running speed.

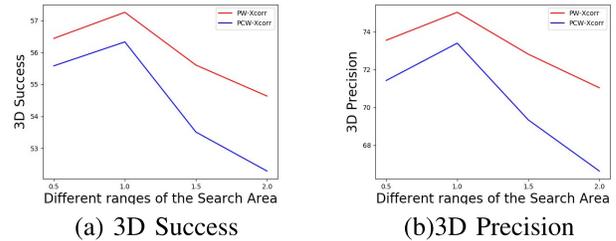

Fig. 8. The curves of three metrics with the distance of search area increasing of car.

*4) Range of Search Area for 3D Car Tracking:* Different search area results in different search point clouds. On one hand, choosing the appropriate search area can properly introduce noise and improve the robustness of the tracking system. On the other hand, the search area can also contain the possible position of the object in the next frame, which is beneficial for object tracking. Therefore, the range of $D$ is important to the tracking results. In this part, we evaluate the performance when using different ranges of $D$, from $0.5m$ to $2m$. Since the values along $Y$ axis (height in the vertical direction) changes little, we do not change the range on $Y$ axis but only on $X$ and $Z$ axis. As Fig. 8 shows, $D = 1.0m$ has the best performance in our proposed methods. We speculate that there are two reasons. First, there are no enough points in search point clouds that represent the target object if $D$ is too small. Second, too large $D$ will introduce too many noisy points from other objects into the search point clouds, which results in the degradation of the tracking performance.

*5) Range of Weight Coefficient for 3D Car Tracking:* Different weight coefficient $\lambda$ also leads to different ratio of classification and regression in loss function. In this part, we evaluate the tracking performance of PW-Xcorr tracker by using different weight coefficients $\lambda$ from 0.1 to 20, and find the most suitable weight coefficient value. Table III reports the car tracking results of the different weight coefficient on KITTI. As it shows, $\lambda = 10$ results in the best overall tracking performance in our method.

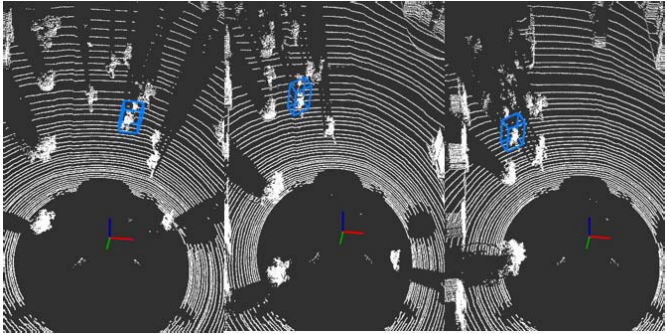

Fig. 9. The example of pedestrian tracking in a crowded scene where the target person ID is 38 in sequence 19 of KITTI dataset. The blue box is the target object. In this scenario, approximately 14 pedestrians are moving in an area of $20m^2$.

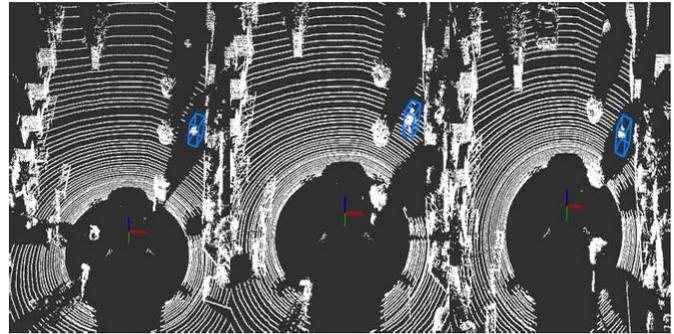

Fig. 10. The example of pedestrian tracking in an occluded scene where the target person ID is 1 in sequence 19 of KITTI dataset. The blue box is the target object. When the relative distance is $10.2m$, the occlusion rate is 68%. When the relative distance is $5.8m$, the occlusion rate is 54%.

### E. 3D Single Non-Rigid Object (Pedestrian) Tracking on KITTI

Pedestrian and car tracking are quite different. First, a car belongs to rigid object and its point cloud feature is stable, such as the square vehicle body, round tires and other intrinsic features of cars. However, the pedestrian's point cloud feature representation is unstable, such as different arm or leg posture, resulting in various feature representation. Second, a pedestrian tends to be much smaller than cars in the real world. Because of the size difference between pedestrians and cars, the point cloud representation of pedestrians is sparser than cars at the same detection distance. For those reasons, it is more challenging to extract stable features and track non-rigid objects (pedestrian) than rigid objects (car).

The previous pedestrian tracking work usually depends on the trained classifier [57], [58] to detect the object if their shape have changed, but did not focus on the tracker to solve this problem. However, our method could track pedestrian in an end-to-end manner and take into consideration of the shape deformations that occurs when a person is walking. In order to evaluate the performance of our method for non-rigid object (pedestrian), we carried out 5 different experiments on KITTI dataset. In the first experiment, we qualitatively analyze the performance of our method in several challenging scenes. The challenging scenes include crowded scenes, occluded scenes and rotating scenes, where rotating scenes are scenes in which target object is rotating around the LIDAR sensor. In the second experiment, we quantitatively evaluate the performance of our method. In the third experiment, we compare the performance differences between our cross-correlation modules (PCW-Xcorr, PW-Xcorr) and other commonly used cross-correlation modules. In the fourth experiment, we test the effect of the search range on the performance of the method. In the fifth experiment, we analyze the effect of different weight coefficient on the tracking performance.

*1) 3D Pedestrian Tracking in Challenging Scenes:* The tracking performance of the proposed method for pedestrian tracking on the KITTI dataset is shown in Fig. 9, Fig. 10 and

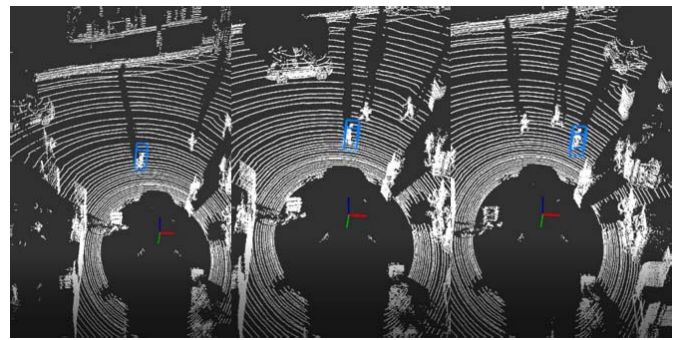

Fig. 11. The example of pedestrian tracking in a rotating scene where the target person ID is 86 in sequence 19 of KITTI dataset. The blue box is the target object. In this scenario, the angular velocity of the pedestrian is approximately $3.7°/s$.

Fig. 11, where the blue bounding box is the target pedestrian. In Fig. 9, the proposed method can track a pedestrian in a crowded scene. In this scenario, approximately 14 pedestrians were moving in an area of $20m^2$, so the search area will introduce many noise point cloud from other pedestrians. As shown in Fig. 10, when the pedestrian is at a distance of $10.2m$, the number of points in the target pedestrian point cloud is only about 102. At the same distance, the number of points in the point clouds for unoccluded pedestrian is generally 320, meaning the occlusion rate is about 68%. When the relative distance is $5.8m$, the number of points in the target pedestrian point cloud is 274. At the same distance, the number of points in the point clouds for unoccluded pedestrian is generally 600, meaning the occlusion rate is about 54%. Fortunately, our method could still work well and cope with the surrounding disturbance effectively. As shown in Fig. 10, our methods still have good tracking performance even in an occluded scene (watch our experiment video for more details). Besides, in Fig. 11, it shows that our method could track target pedestrian in a rotating scene. In such kind of scene, it is more challenging for robust tracking since the relative movement of the target between consecutive frames are usually bigger than other scenes. In this experiment, the angular velocity of the target pedestrian is approximately $3.7°/s$. Experiment results show that our method could deal with such kind of scenario robustly.

TABLE IV
EVALUATION RESULTS OF 3D PEDESTRIAN TRACKING ON KITTI

| Module | 3D Success | 3D Precision | BEV Success | BEV Precision | FPS |
|---|---|---|---|---|---|
| 3D Siamese-2D proposal [11] | 17.89 | 47.81 | - | - | - |
| 3D Siamese-KF [11] | 18.24 | 37.78 | - | - | - |
| P2B [12] | 28.7 | 49.68 | - | - | **46.2** |
| Ours(PCW-Xcorr) | 32.21 | 52.14 | 45.27 | 68.32 | 16.1 |
| Ours(PW-Xcorr) | **33.95** | **53.48** | **53.74** | **70.24** | 18.3 |

TABLE V
PERFORMANCE FOR DIFFERENT MODULES OF 3D PEDESTRIAN ON KITTI

| Module | 3D Success | 3D Precision | BEV Success | BEV Precision | FPS |
|---|---|---|---|---|---|
| No similarity Module | 19.32 | 34.24 | 30.85 | 53.42 | **19.8** |
| DW-Xcorr | 29.74 | 47.52 | 49.36 | 66.13 | 17.9 |
| Cosine Similarity | 29.02 | 49.88 | 43.73 | 67.86 | 17.5 |
| Euclidean Similarity | 33.69 | 48.17 | 50.61 | 68.95 | 16.8 |
| Ours(PCW-Xcorr) | 32.21 | 52.14 | 45.27 | 68.32 | 16.1 |
| Ours(PW-Xcorr) | **33.95** | **53.48** | **53.74** | **70.24** | 18.3 |

*2) Evaluation of 3D Pedestrian Tracking:* As shown in Table IV, our proposed methods both surpass the 3D Siamese-KF [9], 3D Siamese-2D proposal(top-16) [11] and P2B [12] in all metrics. Meanwhile, compared to 3D Siamese-2D proposal [11] method, our proposed tracker with PW-Xcorr performs better by a large margin of 16.06% and 5.67% on 3D Success and 3D Precision separately. Additionally, our proposed methods could run in real time. The proposed PCW-XCorr and PW-XCorr trackers achieve 16.1 FPS and 18.3 FPS respectively.

*3) Cross Correlation Module Ablation Studies for 3D Pedestrian Tracking:* Here, we compare them with the most widely used cosine and euclidean similarity methods for non-rigid (pedestrian) tracking. Additionally, we also compare them with the depth-wise cross correlation (DW-XCorr) [28]. Meanwhile, we test whether the removal of the cross correlation module affects the non-rigid object tracking performance. The results on KITTI are shown in Table V. The proposed module PW-XCorr outperforms the others in three out of four metrics. Similar to car tracking, the two proposed modules outperform the other three modules. Meanwhile, compared to PW-XCorr module, the network without similarity module performs worse by a large margin of 14.63%, 19.24%, 22.89% and 16.82% on 3D/BEV Success and Precision respectively. Additionally, PW-Xcorr performs better than DW-XCorr [28] by 4.21%, 5.96%, 4.38% and 4.11% on 3D/BEV Success and Precision respectively. The results also show that our proposed modules are better than DW-XCorr module for non-rigid object tracking. However, compared to the vehicle tracking, the overall performance of the pedestrian tracking has decreased. We argue that the performance decline is due to the non-rigid characteristics of a person. When a person is walking, the change of limbs will result in the difference of the extracted point cloud feature.

*4) Range of Search Area for 3D Pedestrian Tracking:* Unlike car tracking, considering the non-rigid characteristics of a person and the difference of moving range of a person versus a vehicle in consecutive frames, we also need to evaluate the performance when using different ranges of $D$, from $0.3m$ to

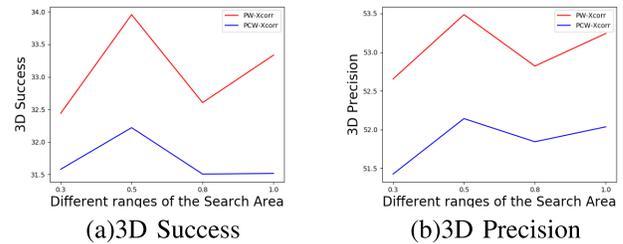

(a) 3D Success  (b) 3D Precision

Fig. 12. The curves of three metrics with the distance of search area increasing of pedestrian.

$1m$. As Fig. 12 shows, $D = 0.5m$ has the best performance in our proposed method. Based on the above results, we speculate that there are two reasons. Similar to vehicle tracking, too small search area will not have enough search point clouds, while too large search area will introduce too much noises. Additionally, we think this is because people have a smaller moving range when walking, while our search range has expanded by 0.5 meters in all directions between the two frames. Therefore, the search box could not only contain point cloud of the target in adjacent frames, but also bring in appropriate noisy points to improve the robustness of our method.

*5) Range of Weight Coefficients for 3D Person Tracking:* As mentioned above, compared with rigid object tracking (car), non-rigid object (pedestrian) tracking is more challenging. Due to the difference between rigid body tracking and non-rigid body tracking, we still need to test the performance of various weight coefficient on the tracking results. As Table VI shown, the pedestrian tracking performance is better than other coefficient value in three out of five metrics when $\lambda = 10$. We can see that the coefficient for car tracking is also applicable to pedestrian tracking.

### F. Extensive Comparisons on KITTI

In this part, we further compared our method with P2B and SC3D on Pedestrian, Van, and Cyclist(Table VII). All methods adopt the same parameter setting. The template point cloud is initialized with the point cloud of the first frame GT and is updated by fusing the point cloud of first frame GT with previous result. The search point cloud is updated based on the point cloud of previous result, which better suits the requirement of real scenes. Our method outperforms P2B and SC3D by ∼4% and ∼15% on average. Generally, our method has better performance in the mean 3D Success/Precision in various target object categories. However, the performance of our method is decreased for Van and Cyclist. And P2B also has the same problems. We think the reason is that our method relies on more data for more accurate similarity calculation and stable feature learning.

TABLE VI
PERSON TRACKING PERFORMANCE FOR DIFFERENT WEIGHTS ON KITTI

| weight($\lambda$) | 3D Success | 3D Precision | BEV Success | BEV Precision |
|---|---|---|---|---|
| 0.1 | 31.42 | 49.22 | 47.06 | 69.52 |
| 0.5 | 31.13 | 52.65 | 44.92 | 68.43 |
| 1.0 | 30.13 | 49.84 | 46.83 | 68.67 |
| 5.0 | 33.22 | 53.16 | 50.20 | 69.18 |
| 10.0 | **33.95** | 53.48 | **53.74** | **70.24** |
| 20.0 | 33.94 | **54.82** | 47.69 | 69.30 |

TABLE VII
DETAILS OF EXTENSIVE COMPARISONS WITH SC3D AND P2B ON KITTI

| | Method | Car | Pedestrian | Van | Cyclist | Mean |
|---|---|---|---|---|---|---|
| | Frame Number | 6424 | 6088 | 1248 | 308 | 14068 |
| 3D Success | SC3D-KF [9] | 41.3 | 18.2 | 40.4 | **41.5** | 31.2 |
| | SC3D-EX [9] | 21.2 | 8.1 | 25.2 | 11.1 | 15.7 |
| | P2B [12] | 56.2 | 28.7 | 40.8 | 32.1 | 42.4 |
| | Ours(PCW-Xcorr) | 57.56 | 33.54 | 44.32 | 35.92 | 45.52 |
| | Ours(PW-Xcorr) | **58.22** | **35.23** | **45.67** | 36.16 | **46.67** |
| 3D Precision | SC3D-KF [9] | 57.9 | 37.8 | 47.0 | **70.4** | 48.5 |
| | SC3D-EX [9] | 24.8 | 14.2 | 28.9 | 14.8 | 20.35 |
| | P2B [12] | 72.8 | 49.6 | 48.4 | 44.7 | 60.0 |
| | Ours(PCW-Xcorr) | 74.40 | 54.15 | 51.46 | 48.17 | 63.03 |
| | Ours(PW-Xcorr) | **76.24** | **56.22** | **52.85** | 49.03 | **64.91** |

TABLE VIII
DIFFERENT WAYS FOR TEMPLATE GENERATION AND SIZE INFORMATION GENERATION

| Source of template points | Source of size information | Method | 3D Success | 3D Precision | FPS |
|---|---|---|---|---|---|
| the First GT | the GT whl | Ours(PW-Xcorr) | **57.25** | **75.03** | **20.8** |
| | the predicted whl | Ours(PW-Xcorr) | 51.09 | 63.96 | 20.8 |
| the First GT & Previous | the GT whl | Ours(PW-Xcorr) | 58.22 | 76.24 | 11.3 |
| | the predicted whl | Ours(PW-Xcorr) | 52.35 | 64.82 | 11.3 |

[1] "the First GT & Previous" denotes "The first GT and Previous result". "the GT whl" denotes "using w,h,l with GT". "the predicted whl" denotes "using w,h,l with our method predicted".

### G. Ablation Study for Template Generation and Different Source of Size Information

In this ablation study, we analyze the effect of different template point cloud generation methods and different size information generation methods on tracking performance. For template generation, we follow the 2D visual tracking methods [22], [26], [27], [30] to freeze and no longer update the template point cloud with the first GT in our method. However, inspired from P2B [12], we can update template by concatenating the point cloud within the first GT and previous result. Here, we reported results with two settings for template generation: the first GT(our default setting) and the fusion of the first GT and previous result's point clouds(P2B default setting). Results in Table VIII show that the performance of our method only slightly improved ∼1% after updating the template point cloud. However, the running speed of our method will decrease from 20.8 FPS to 11.3 FPS. We think that there may be two reasons. First, most targets move from near to far due to the characteristics of KITTI dataset. Therefore, the number of the first GT's point clouds is more, and the feature representation of target is rich enough, so it is no longer necessary to fuse the previous results. Second, for some bad tracking results, if they are fused into the template point cloud, too much noise will be introduced into the template, which will affect the performance of our method. Meanwhile, when we update template, the running time of our method will drop from 20 FPS to 11 FPS. Because we need to recalculate the template features in every frame. Besides, P2B and SC3D use the object's GT size information(w,h,l) because they did not predict the size information of the target. However, our method predicts the size information of the target, because this information is very important in real-scene applications. When we use the size information of the GT, the performance of our method improves ∼6%. This is because the regression target of our model during training is the mean size value of all vehicles. Therefore, the predicted size information is not accurate enough. However, our method is more suitable for real-scene applications.

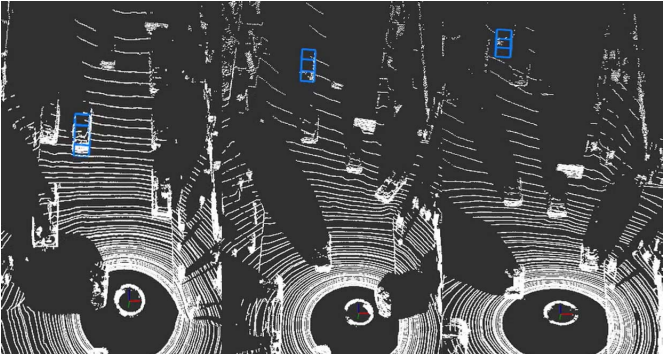

Fig. 13. The example of car tracking in an extremely sparse scene where the target vehicle ID is 21 in scenario 11 of H3D dataset. The blue box is the target object. In this scenario, when the relative distance is about 49$m$, the number of points for target car point clouds is only about 12. When the relative distance is about 39$m$, the number of points for target car point clouds is about 29.

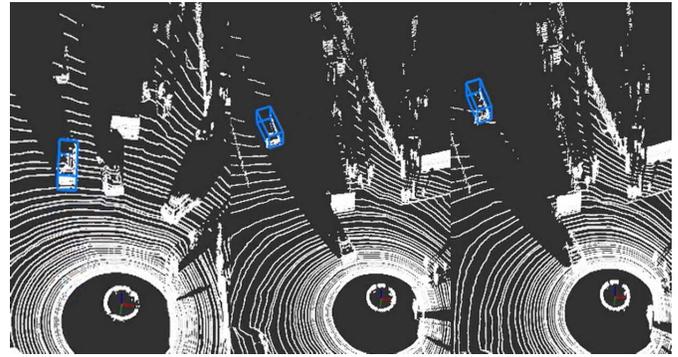

Fig. 14. The example of car tracking in an occluded scene where the target car ID is 26 in scenario 11 of H3D dataset. The blue box is the target object. In this scenario, when the relative distance is about 28$m$, the occlusion rate is 59%. When the relative distance is about 23$m$, the occlusion rate is 56%.

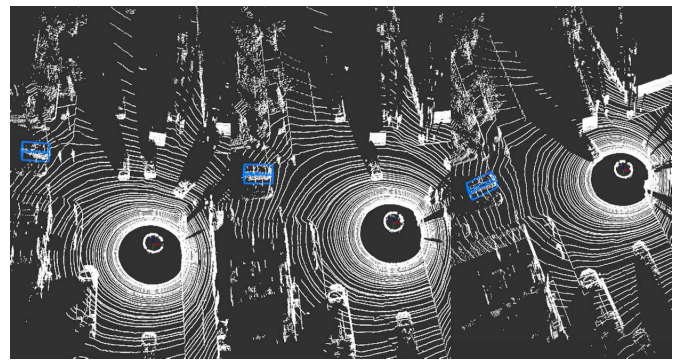

Fig. 15. The example of car tracking in a turning scene where the target car ID is 89 in scenario 11 of H3D dataset. The blue box is the target object. In this scenario, the angular velocity of LIDAR is approximately 9.7°/$s$.

### H. 3D Single Object Tracking on H3D

As mentioned before, H3D is collected from crowded urban environments and is more complex and challenging than the simple scenes of KITTI dataset. Unlike 3D object annotations are labeled in the frontal view on KITTI, 3D objects in H3D dataset are labeled in the full surround scene. Therefore, this dataset is very different from KITTI dataset. Here, we use H3D dataset to test our approach with models trained on KITTI dataset to verify the generalization ability of the proposed method. We carried out 2 different experiments on H3D dataset. In the first experiment, we qualitatively analyze the performance of our method in several challenging scenes for rigid object (car) and non-rigid object (pedestrian) tracking. In the second experiment, we quantitatively evaluate the performance of our method. The challenging scenes include extremely sparse scenes, occluded scenes and turning scenes. In sparse and occluded scenes, because the point cloud of the target is extremely sparse or occluded by other objects, which would lead to incomplete point cloud representation of the target, thus bringing great challenges for network to discriminate and track. In turning scenes, it is more challenging for robust tracking since the relative movement of the target between consecutive frames are usually bigger than other scenes. Therefore, we chose these scenarios to test the robustness of our method.

*1) 3D Car and Pedestrian Tracking in Challenging Scenes:* The qualitative tracking performance of our method for car tracking on the H3D dataset is shown in Fig. 13, Fig. 14 and Fig. 15, where the blue bounding box is the target car. For extremely sparse scenes, as shown in Fig. 13, the number of points for target car point clouds is only 12 when the relative distance is about 49$m$. Even the target car has extremely sparse point clouds, our method could still track it robustly. For occluded scenes, as shown in Fig. 14, most of the point clouds of the target car are occluded by other cars, which would lead to incomplete point cloud representation for the target, thus making it more difficult to match with the template point clouds. In Fig. 14, when the relative distance is 28$m$, the number of points in the target car point cloud is 127.

At the same distance, the number of points in the point clouds for unoccluded car is generally 310, meaning the occlusion rate is about 59%. However, our method could still effectively track the target car even in such a scene. For turning scenes, as shown in In Fig. 15, the angular velocity of LIDAR is approximately 9.7°/$s$. Our method could track the target car in a turning scene successfully. The qualitative tracking performance of our method for pedestrian tracking on the H3D dataset is shown in Fig. 16, Fig. 17 and Fig. 18, where the blue bounding box is the target pedestrian. For extremely sparse scenes, as shown in Fig. 16, the number of points for target pedestrian point clouds is only 25 when the relative distance is about 38$m$. Our method could still track a pedestrian in such sparse scenes. For occluded scenes, as shown in Fig. 17, when the pedestrian is at a distance of 20$m$, the number of points for target pedestrian point cloud is only 21. At the same distance, the number of points for unoccluded pedestrian is generally 104, meaning the occlusion rate is about 80%. However, our method could effectively track the target pedestrian robustly. For turning scenes, as shown in In Fig. 18, the angular velocity of LIDAR is approximately 9.7°/$s$. Our method could track target pedestrian in a turning scene. Experiment results show that the model trained on KITTI dataset could track targets

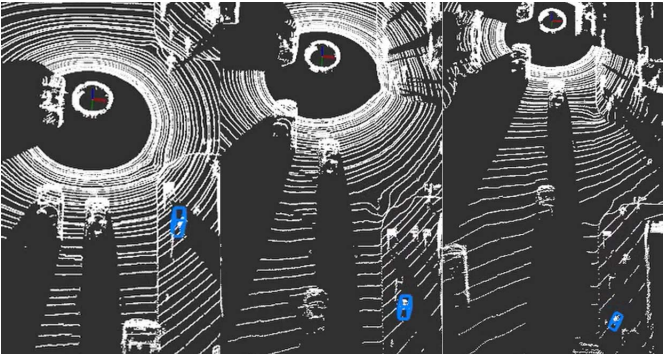

Fig. 16. The example of pedestrian tracking in an extremely sparse scene where the target pedestrian ID is 14 in scenario 11 of H3D dataset. The blue box is the target object. In this scenario, when the relative distance is about 38$m$, the number of points for target pedestrian is only about 25. When the relative distance is about 30$m$, the number of points for target pedestrian is about 37.

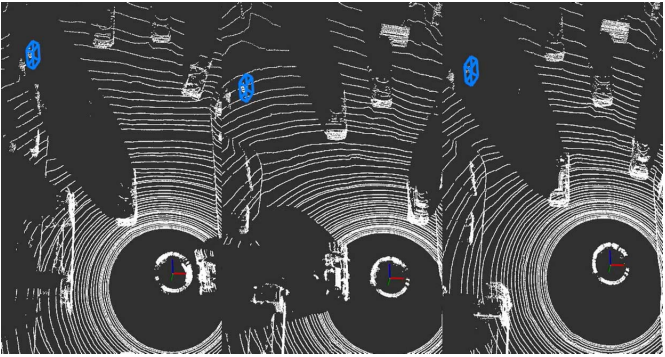

Fig. 17. The example of pedestrian tracking in an occluded scene where the target person ID is 87 in scenario 11 of H3D dataset. In this scenario, when the relative distance is about 20$m$, the occlusion rate is 80%.

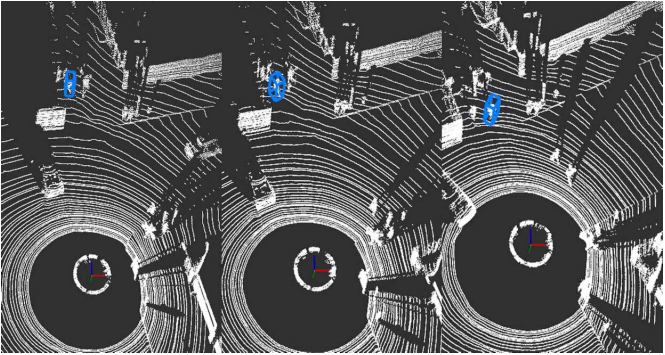

Fig. 18. The example of pedestrian tracking in a turning scene where the target person ID is 107 in scenario 11 of H3D dataset. The blue box is the target object. In this scenario, the angular velocity of LIDAR is approximately 9.7$°/s$.

robustly in the above challenging scenarios, showing the good robustness.

*2) Evaluation of 3D Car and Pedestrian Tracking on H3D:* For car tracking, as shown in Table IX, since the target object may disappear for a while and the sizes of cars are different in the two datasets, the evaluation results have a decline compared to that on KITTI. For pedestrian tracking, as shown in Table X, even the tracking performance reduces a bit, the proposed methods can still work effectively in an untrained new environment. It should be noted that the network models were trained on KITTI dataset and we did not train it using any sequences from H3D dataset. However, the proposed methods still have good performances. The results show that the proposed methods have good generalization ability on different datasets.

TABLE IX
DETAILS OF EVALUATION RESULTS OF CAR TRACKING ON H3D

| Module | 3D Success | 3D Precision | BEV Success | BEV Precision |
|---|---|---|---|---|
| Ours(PCW-Xcorr) | 41.49 | **54.29** | **57.31** | **65.94** |
| Ours(PW-Xcorr) | **41.98** | 53.70 | 56.07 | 62.68 |

TABLE X
DETAILS OF EVALUATION RESULTS OF PEDESTRIAN TRACKING ON H3D

| Module | 3D Success | 3D Precision | BEV Success | BEV Precision |
|---|---|---|---|---|
| Ours(PCW-Xcorr) | **16.52** | **45.18** | **24.37** | **57.44** |
| Ours(PW-Xcorr) | 15.43 | 44.86 | 23.50 | 56.31 |

TABLE XI
RUNNING TIME OF DIFFERENT METHODS FOR CAR

| Method | pre-process | model-inference | post-process | total |
|---|---|---|---|---|
| SC3D-KF [9] | 418.23ms | 2.13ms | 34.18ms | 454.54ms |
| SC3D-EX [9] | 520.84ms | 2.09ms | 34.31ms | 557.24ms |
| P2B [12] | 7ms | 14.3ms | 0.9ms | 22.2ms |
| **Ours(PCW-Xcorr)** | **0.52ms** | **51.99ms** | **7.39ms** | **59.9ms** |
| **Ours(PW-Xcorr)** | **0.48ms** | **40.27ms** | **7.20ms** | **47.95ms** |

*I. Running Speed*

Here we calculate the average running time of all test frames for car to measure the speed of our method, and compare it with P2B and SC3D. As shown in Table XI, 3D-SiamRPN achieved 20.8FPS with PW-Xcorr module, including 0.48ms for cropping point cloud, 40.27ms for network forward propagation and 7.2ms for post-processing, on a Nvidia 1080ti GPU. P2B and SC3D both used Kalman Filter to generate search space of candidates, and their running speeds are 45 FPS and 2.2 FPS respectively. Notably, we found that our method takes a shorter time(0.5ms) in data pre-process, while P2B(7ms) and SC3D(420ms) both take a longer time because they need to generate potential object proposals from the search space. Especially for SC3D, they generate 147 proposals, resulting in a longer time consuming. Besides, our method takes a longer running time in model inference. Although we use the same backbone network as P2B for point cloud feature extraction, our network is different from P2B. We used Feature Propagation(FP) layer in PointNet++ [14] because it focuses more on the global semantic feature of point cloud, while P2B does not use it. We think this is the reason why our method performs better than P2B. In our method, each part of our network takes the following time during the inference stage: point cloud feature extraction takes 36.3ms, feature fusion takes 1.7ms and RPN network takes 2ms.

## V. Conclusion

In this paper, we proposed 3D-SiamRPN to solve 3D single object real-time tracking problem, which does not rely on the detector and directly obtains the 3D bounding box of the target object by an end-to-end learning manner. Meanwhile, we proposed two types of cross correlation modules for point cloud features. The experimental results on different datasets show that the proposed method is competitive with the state-of-the-art methods in tracking rigid objects and non-rigid objects. Meanwhile, the proposed modules PCW-XCorr and PW-XCorr shows better performances than the other similarity modules in 3D object tracking. Additionally, we demonstrate the good generalization ability of our method by validating the proposed method in a fully new dataset without re-training. In the future, we plan to effectively fuse the good previous tracking results to update the template point cloud and extend our approach to multi-object tracking.

## Acknowledgment

The authors would also like to thank J. Shan, W. Qiao and M. Zhou for their help.

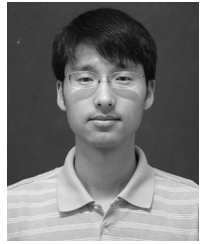

**Zheng Fang** (Member, IEEE) received the B.S. degree in automation and the Ph.D. degree in pattern recognition and intelligent systems from Northeastern University, China, in 2002 and 2006, respectively. He was a Postdoctoral Research Fellow of Carnegie Mellon University from 2013 to 2015. He is currently an Associate Professor with the Faculty of Robot Science and Engineering, Northeastern University. His research interests include visual/laser SLAM, and the perception and autonomous navigation of various mobile robots.

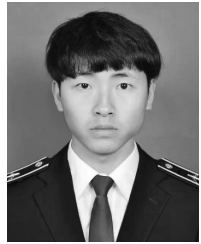

**Sifan Zhou** received the B.S. degree from the Civil Aviation University of China. He is currently pursuing the M.S. degree with Northeastern University. His research interests include deep learning, visual perception, and 3D object tracking.

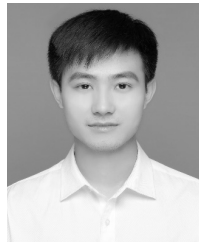

**Yubo Cui** received the B.S. degree in automation and the M.S. degree in robot science and engineering from Northeastern University, China, in 2017 and 2020, respectively. His research interests include deep learning, 3D object tracking, and 3D detection.

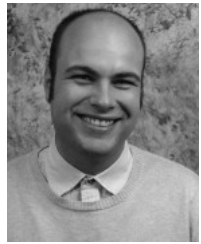

**Sebastian Scherer** (Senior Member, IEEE) received the B.S. degree in computer science, and the M.S. and Ph.D. degrees in robotics from Carnegie Mellon University (CMU), Pittsburgh, PA, USA, in 2004, 2007, and 2010, respectively. He is an Associate Research Professor of the Robotics Institute, CMU. He and his team have demonstrated the fastest and most tested obstacle avoidance on a YamahaRMax in 2006, the first obstacle avoidance for microaerial vehicles in natural environments in 2008, and the first (2010) and fastest (2014) automatic landing zone detection and landing on a full-size helicopter. His research interest includes enabling autonomy for unmanned rotorcraft to operate at low altitude in cluttered environments.